\documentclass[letterpaper, 10 pt, conference]{ieeeconf}  

\usepackage[utf8]{inputenc}
\usepackage{textcomp}
\usepackage{graphicx}
\usepackage[version=4]{mhchem}
\usepackage{siunitx}
\usepackage{longtable,tabularx}
\usepackage{accents}
\usepackage{comment}
\usepackage{mathtools}
\usepackage{subfigure}
\usepackage{caption}
\usepackage{color}
\usepackage{amsmath,amssymb}
\usepackage{booktabs}
\usepackage{tabu}
\usepackage{algorithm}
\usepackage{algpseudocode}
\usepackage{soul}
\usepackage[dvipsnames]{xcolor}
\usepackage{multirow}
\usepackage{enumerate}
\usepackage{svg}
\usepackage{epstopdf}
\usepackage{soul}
\usepackage{arydshln}

\usepackage{tikz}
\usepackage{siunitx} 
\usetikzlibrary{arrows, arrows.meta}
\usetikzlibrary{graphs}
\usetikzlibrary{snakes,backgrounds,patterns,matrix,shapes,fit,calc,shadows,plotmarks} 
\usetikzlibrary{bayesnet}

\usepackage{standalone}

\tikzstyle{target}=[draw,fill=yellow!50,circle,minimum size=16pt,inner sep=0pt]

\tikzstyle{output}=[draw,fill=blue!50,circle,minimum size=16pt,inner sep=0pt]
\tikzstyle{bias}=[draw,fill=gray!50,circle,minimum size=20pt,inner sep=2pt]
\tikzstyle{arrow}=[arrows={{Latex[scale=0.5]}-}, thick]  
\tikzstyle{box}=[rectangle, draw=black!100] 

\tikzset{
    between/.style args={#1 and #2}{
         at = ($(#1)!0.5!(#2)$)
    }
}

\newcommand{\od}[1]{{\color{ForestGreen}OD: #1} }

\IEEEoverridecommandlockouts                              

\title{\LARGE \bf
Conservative Filtering for Heterogeneous Decentralized Data Fusion in Dynamic Robotic Systems*
}

\author{Ofer Dagan$^{1}$ and Nisar R. Ahmed$^{1}$
\thanks{*This work was not supported by any organization}
\thanks{$^{1}$
Ofer Dagan and Nisar R. Ahmed are with the Smead Aerospace Engineering Sciences Department, University of Colorado Boulder, Boulder, CO 80309 USA {\tt\small ofer.dagan@colorado.edu; Nisar.Ahmed@colorado.edu}}
}

\begin{document}

\maketitle
\thispagestyle{empty}
\pagestyle{empty}

\begin{abstract}
This paper presents a method for Bayesian multi-robot peer-to-peer data fusion where any pair of autonomous robots hold non-identical, but overlapping parts of a global joint probability distribution, representing real world inference tasks (e.g., mapping, tracking). It is shown that in dynamic stochastic systems, filtering, which corresponds to marginalization of past variables, results in direct and hidden dependencies between variables not mutually monitored by the robots, which might lead to an overconfident fused estimate. The paper makes both theoretical and practical contributions by providing (i) a rigorous analysis of the origin of the dependencies and and (ii) a conservative filtering algorithm for heterogeneous data fusion in dynamic systems that can be integrated with existing fusion algorithms. This work uses factor graphs as an analysis tool and an inference engine. Each robot in the network maintains a local factor graph and communicates only relevant parts of it (a sub-graph) to its neighboring robot. We discuss the applicability to various multi-robot robotic applications and demonstrate the performance using a multi-robot multi-target tracking simulation, showing that the proposed algorithm produces conservative estimates at each robot.

\end{abstract}

\section{INTRODUCTION}
Consider a team of robots operating in an unknown environment and collaborating on a global joint inference task such as building a map of an unknown environment \cite{schoenberg_distributed_2009}, tracking a set of targets \cite{dagan_heterogeneous_2020} or cooperatively localizing themselves \cite{loefgren_scalable_2019, li_cooperative_2012, zhu_cooperative_2019}. To enable a scalable operation of such teams, a common practice is to distribute the global joint inference task into smaller, overlapping tasks, and allow robots to fuse only data relevant to their local inference task. Here, for example, relevant data might be an overlapping part of the map; subset of targets or a robot's immediate neighboring vehicles, respectively. Assuming each robot is an independent entity that needs to send and locally fuse messages, the examples above describe different instances of \emph{heterogeneous} fusion \cite{dagan_heterogeneous_2020}.     

There are several approaches for heterogeneous multi-robot cooperation, such as consensus \cite{ sun_scaling_2018}, or by using a server/cloud computing \cite{kia_server-assisted_2018, schmuck_ccm-slam_2019}. This paper explores a peer-to-peer Bayesian heterogeneous decentralized data fusion (DDF) approach, as it has advantages in robustness and flexibility. Briefly, in DDF robots communicate their current posterior probability distribution function (pdf) based on their locally available data \cite{chong_distributed_1983}.   

In previous work \cite{dagan_heterogeneous_2020}, \cite{dagan_exact_2021} we define peer-to-peer Bayesian heterogeneous DDF as the process of fusing two non-equal, but overlapping, pdfs.\footnote{To be precise, we differentiate between the more general problem of forming a posterior pdf over a set of random variables, which might describe a random state vector, and the Bayesian point estimate of those variables, given for example by finding the minimum squared error (MMSE) of that pdf.} One of the main challenges of Bayesian DDF and more acutely of heterogeneous DDF is accounting for dependencies in the data gathered by the different robots. Incorrect treatment of these dependencies might lead to `double counting' data more than once, which results in an overconfident estimate. For the case of homogeneous DDF, i.e. where the communicated and posterior pdfs describe the same set of random variables, it is common to either (i) explicitly track dependencies, which can be done by keeping a pedigree of the incoming data \cite{martin_distributed_2005} or by adding a channel filter (CF) \cite{grime_data_1994} for example; or (ii) implicitly account for dependencies by discounting data using covariance intersection (CI) \cite{julier_non-divergent_1997} or with the geometric mean density (GMD) \cite{bailey_conservative_2012}, for example. Notice that these methods are not application-specific and used to solve a variety of problems. On the other hand, in the case of heterogeneous fusion, solutions to account for the data dependencies have so far been rather application-specific. 

In \cite{cunningham_ddf-sam_2013} a multi-robot SLAM problem is solved, where dynamic robots share static variables (the landmarks describing the unknown map) and the solution is given for a smoothing problem, i.e. robot's local position variables are augmented. Thus this algorithm does not solve the filtering or fixed-lag smoothing problem, which are relevant for target tracking and cooperative localization applications for example. In cooperative localization (CL), robots take relative measurements to their neighboring robots, which directly couples the robots' states. To the best of our knowledge, in current approaches, robots either explicitly keep track of the dependencies of all \cite{luft_recursive_2016} or groups \cite{li_cooperative_2012} of vehicles in the network, or implicitly account for dependencies with CI to approximate the joint coupled covariance by its block diagonal elements and then marginalize out the other robot \cite{kia_cooperative_2016, zhu_cooperative_2019}. While these approaches work for cooperative localization, they might not adapt well for other cooperative applications, e.g., target tracking or SLAM. Thus there is a need to look at heterogeneous fusion for robotics as a wider, non-application specific problem, and develop the necessary analysis tools and algorithms. For that reason, previous work \cite{dagan_exact_2021} aimed at gaining insight into the heterogeneous fusion problem. 

In \cite{dagan_heterogeneous_2020, dagan_exact_2021} it is shown that in order for the fused posterior to be conservative (doesn't underestimate the uncertainty) over all the robot's variables, conditional independence has to be maintained between non-mutual variables, given the common variables, e.g., in multi-robot SLAM, non-mutual position variables are independent given common landmark variables. An algorithm, named the \emph{Heterogeneous State Channel Filter} (HS-CF) was developed, and shown to be conservative for static problems and for dynamic problems when a smoothing approach was applied. However, for a filtering scenario, the algorithm was slightly overconfident (non-conservative). It was hypothesized that marginalization of past variables (states) in the filtering stage resulted dependencies between non-mutual variables.     

The goal of this work is to understand the nature of the  dependencies between non-mutual variables resulting from filtering in heterogeneous DDF applications. To explicitly track and account for dependencies between common variables, the CF framework is used, which requires the communication graph between robots to be an undirected a-cyclic graph, so data can flow only in one route (i.e., can not `circle' back). 
Factor graphs, representing the local robot's pdf are analysed to gain insight into the structure of the problem and the cause of dependencies between non-mutual variables. While this paper is motivated by a search for rigorous understanding of the heterogeneous fusion problem, to develop new theory and thus focus less on a specific application, the presented approach can be readily applied to different robotic applications. 

The major contributions of this paper are twofold: theoretical and practical. From a theoretical point of view, factor graph analysis i) reveals `hidden variable dependency dynamics' for Bayesian heterogeneous DDF and ii) sheds light on the interplay between groups of common variables, showing that preserving conditional independence structure through the filtering stage is key to helping ensure conservativeness. From a practical point of view, i) an algorithm for conservative filtering that can be integrated with other fusion algorithms is developed and ii) multi-robot target tracking simulation verifies that the presented approach overcomes previous difficulties for dynamic variables.

The rest of the paper is organized as follows: Sec. \ref{Sec:background} gives the necessary background on heterogeneous DDF problems and factor graphs; Sec. \ref{Sec:prob_statement} defines the problem of conservative filtering in the context of heterogeneous DDF and Sec. \ref{Sec:cons_filtering} details the analysis, presents the suggested solution and algorithm. Simulation results, discussion and conclusions are given in Sec. \ref{Sec:sim} and Sec. \ref{Sec:conclusions}.

\section{BACKGROUND}
\label{Sec:background}
Consider a set $N_r$ of $n_r$ autonomous robots jointly tasked with monitoring a global set of random variables $V$. These random variables can be static, dynamic or even parameters of the distribution, and can describe, for example, the random position states of the $n_r$ robots and $n_t$ targets tracked by the robots. Since the size $V$ scales with the number of robots and targets and might become quite heavy to process locally by each robot, there is interest in distributing it and tasking each robot $i\in N_r$ with only a subset of random variables $\chi^i\subset V$. To allow for  collaboration between robots we take a Bayesian DDF approach. 

In Bayesian DDF, each robot $i$ is an independent entity, collecting data over a set of random variables of interest $\chi^i$ from its local sensors and/or via communication with neighboring robots. The robots can then individually (locally) infer the posterior distribution over their random variables of interest via Bayes' rule $p(\chi^i|Z^i)=p(\chi^i)p(Z^i|\chi^i)$, where $Z^i$ is the locally available data at robot $i$ and $p(\chi^i)$ is the prior distribution. Note that if two robots $i$ and $j$ hold distributions over overlapping sets of variables, i.e., $\chi^i\cap \chi^j\neq\varnothing$ and $\chi^i\setminus \chi^j\neq\varnothing$, where `$\setminus$' is the set exclusion operation, \cite{dagan_exact_2021} defines the resulting fusion process as an instance of \emph{heterogeneous} fusion.

\subsection{Heterogeneous Bayesian Decentralized Data Fusion}
In heterogeneous DDF a robot's set of random variables of interest $\chi^i$ can be divided into: (i) a set of variables it has in common with at least one of its set of neighbors $N_r^i$, such that $\chi^i_C=\bigcup_{j\in N_r^i}^{}\chi^{ij}_C$, where $\chi^{ij}_C$ are the variables common to robots $i$ and $j$, and (ii) a set of local variables $\chi^i_L$ which are not monitored by any other robot in the network. Robot $i$'s set of variables is thus $\chi^i=\chi^i_L\bigcup \chi^i_C$.

Assume that non-mutual variables to robots $i$ and $j$, defined by $\chi^{i\setminus j}=\chi^i_L\cup\{\chi^i_C\setminus \chi^{ij}_C \}$, are independent given the common variables, i.e., $\chi^{i\setminus j}\perp \chi^{j\setminus i}|\chi^{ij}_C$. Then using a distributed version of Bayes' rule, \cite{dagan_heterogeneous_2020} shows that the \emph{Heterogeneous State} (HS) fusion rule at robot $i$ is,
\begin{equation}
    \begin{split}
        p_f(\chi^i|Z^{i,+}_k)\propto
        \frac{p(\chi^{ij}_C|Z^{i,-}_k)p(\chi^{ij}_C|Z^{j,-}_k)}{p^{ij}_c(\chi^{ij}_C|Z^{i,-}_k \cap Z^{j,-}_k)}  
         \cdot p(\chi^{i\backslash j}|\chi^{ij}_C,Z^{i,-}_k).
    \end{split}
    \label{eq:Heterogeneous_fusion}
\end{equation}
Here $Z^{i,-}_k$ and $Z^{i,+}_k$ are the data at time step $k$, available at robot $i$ before and after fusion, respectively, and it is assumed that all measurements acquired by each robot are conditionally independent given the random variable of interest.

There are two key points concerning the above fusion rule. First, the validity of the equation is based on the assumption that non-mutual variables are conditionally independent given common variables between the communicating robots. However, when the variables of interest are dynamic and are marginalized out successively over time, dependencies between non-mutual variables arise and the fusion rule (\ref{eq:Heterogeneous_fusion}) is no longer valid. This problem is at the core of this paper and is demonstrated, analyzed and solved in the next sections. The second point is the importance of accounting for the posterior distribution over the common variables, given the data common to both robots, $p^{ij}_c(\chi^{ij}_C|Z^{i,-}_k \cap Z^{j,-}_k)$, in the denominator of (\ref{eq:Heterogeneous_fusion}). 

There are several methods to approximate this distribution (e.g., CI \cite{julier_non-divergent_1997}, GMD \cite{bailey_conservative_2012}) when the dependencies in the data held by two robots are unknown. However, the focus of current work is understanding the nature and origin of these dependencies. Thus we choose to use a CF \cite{grime_data_1994} to try to explicitly track the data common and dependencies between the robots.   

\subsection{The Channel Filter}
\label{subsec:CF}
The Channel filter (CF) was suggested in \cite{grime_data_1994} as a method to track dependencies in the data in a network of robots. The main idea is to add a filter on the communication channel between any two communicating robots. It was shown \cite{grime_data_1994}, that for homogeneous fusion, if: (i) data does not circle back to its sender, i.e., the communication graph is undirected and a-cyclic (e.g., tree or chain); (ii) there is full rate communication (i.e., communication at every time step, without delays); (iii) incoming data is processed sequentially and (iv) assuming linear dynamic and measurement models and additive white Gaussian noise (AWGN), then the CF allows for each robot to recover the optimal centralized state estimate.   

For heterogeneous fusion, since robots only reason over parts of the global set of random variables, they cannot achieve the optimal estimate. Nevertheless, in \cite{dagan_heterogeneous_2020}, \cite{dagan_exact_2021} an extended version of the original homogeneous CF is developed - the \emph{Heterogeneous State CF} (HS-CF). In the HS-CF algorithm each CF maintains a distribution over only the common variables between any two communicating robots (as opposed to the full set of variables in the original CF), given the common data. Thus the distribution $p^{ij}_c(\chi^{ij}_C|Z^{i,-}_k \cap Z^{j,-}_k)$ is explicitly tracked and can be accounted for when fusing according to (\ref{eq:Heterogeneous_fusion}). 

The HS-CF was shown (\cite{dagan_exact_2021}) to be conservative for static problems and for dynamic problems with a smoothing approach. However, for a filtering scenario, it resulted in a slightly overconfident state minimum mean squared error (MMSE) estimates. These results revealed a gap in both the understanding of the dependencies resulting from marginalizing out past variables (filtering) and in the tools available to analyze heterogeneous DDF, more specifically the conditional independence structure of such problems. 

\subsection{Factor Graphs}
In recent years factor graphs \cite{frey_factor_1997} have been used to study and solve a variety of robotic applications \cite{dellaert_factor_2021}. Factor graphs are arguably the most general framework to analyze and express conditional independence, as such they directly express the sparse structure of decentralized problems.

A factor graph is an undirected bipartite graph $\mathcal{F}=(U,V,E)$ that represents a function, proportional to the joint pdf over all random variable nodes $v_m\in V$, and factorized into smaller functions given by the factor nodes $f_l\in U$. An edge $e_{lm}\in E$ in the graph only connects a factor node \emph{l} to a variable node \emph{m}.
The joint distribution over the graph is then proportional to the global function $f(V)$:
\begin{equation}
    p(V)\propto f(V)=\prod_{l}f_l(V_l),
    \label{eq:factorization}
\end{equation}
where $f_l(V_l)$ is a function of only those variables $v_m\in V_l$ connected to the factor \emph{l}.

In \cite{dagan_factor_2021} a factor graph based DDF framework, \emph{FG-DDF}, was suggested and demonstrated on a static heterogeneous fusion problem. Here, we further use factor graphs to analyze conservative filtering in dynamic problems.

\section{PROBLEM STATEMENT}
\label{Sec:prob_statement}
Consider a robot $i$ from the network of $n_r$ robots described in Sec. \ref{Sec:background}. Without loss of generality, assume that this robot has two neighbors $j$ and $m$ and the three robots are part of an undirected a-cyclic communication topology, such that $j-i-m$ (i.e., $j$ does not have a way to receive data from $m$ other than through $i$). Robot $i$ exchanges data over its sets of common variables $\chi^{ij}_C$ and $\chi^{im}_C$ with robots $j$ and $m$, respectively, via the HS-CF fusion rule (\ref{eq:Heterogeneous_fusion}). Suppose that $V$, the global set of variables monitored by the $n_r$ robots, describes some systems' random state vector, constructed out of dynamic and static states.

Every robot $i$ gathers data and updates its prior pdf in two ways: (i) by independent sensor measurements $y^{i,l}_{k}\in Y^i_k$, described by the conditional likelihood $p(Y^i_k|\chi^i_k)=\prod_{l}p(y^{i,l}_{k}|\chi^{i,l}_{k})$ and updated using Bayes' rule, where $\chi^{i,l}_{k}$ is the subset $l$ of states, measured by the $y^{i,l}_{k}$ measurement, taken by robot $i$ at time step $k$; (ii) peer-to-peer heterogeneous fusion of data $Z^{j,-}_k$ from a neighboring robot $j\in N_r^i$, updated via (\ref{eq:Heterogeneous_fusion}). 

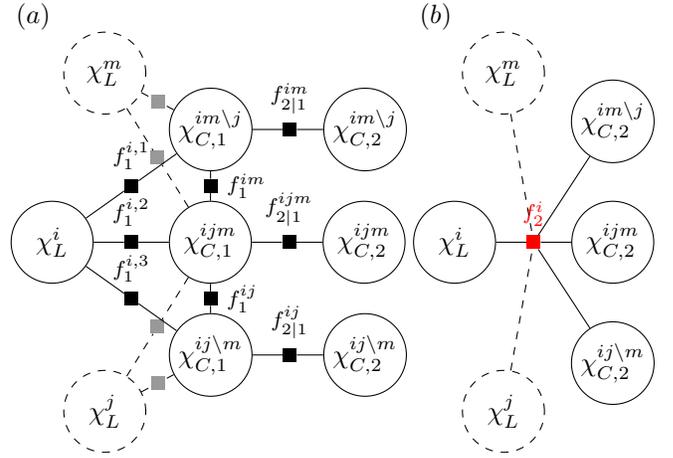
\begin{figure}[tb]
\scalebox{0.93}{}{%
\begin{tikzpicture}[ new set=import nodes]
 \begin{scope}[nodes={set=import nodes}]
      
      \node (a) at (-0.35,3) {$(a)$};
      \node (x_Li)[latent, minimum size=31pt] at (-0.1,0) {$\chi^i_L$};
      \node (x_Lm)[latent,dashed, minimum size=31pt] at (0.6,2.25) {$\chi^m_L$};
      \node (x_Lj)[latent,dashed, minimum size=31pt] at (0.6,-2.25) {$\chi^j_L$};
      \node [latent, right=of x_Li, yshift=+1.5cm, minimum size=31pt] (x_im) {$\chi^{im\setminus j}_{C,1}$};
      \node [latent, right=of x_Li, yshift=0, minimum size=31pt] (x_ijm) {$\chi^{ijm}_{C,1}$};
      \node [latent, right=of x_Li, yshift=-1.5cm, minimum size=31pt] (x_ij) {$\chi^{ij\setminus m}_{C,1}$};
      \node [factor, between=x_Li and x_im,label=$f^{i,1}_{1}$ ] (yi11) {};
      \node [factor, between=x_Li and x_ijm,label=$f^{i,2}_{1}$ ] (yi21) {};
      \node [factor, between=x_Li and x_ij,label=$f^{i,3}_{1}$ ] (yi31) {};
     
      \node [factor, between=x_im and x_ijm ,label=right:$f^{im}_1$] (f_im) {};
      \node [factor, between=x_ij and x_ijm ,label=right:$f^{ij}_1$] (f_ij) {};
     
     \node [factor, right=of x_im,  label={$f^{im}_{2|1}$}] (f2_im) {};
      \node [latent, right=of f2_im, xshift=-0.65cm, minimum size=31pt] (x2_im) {$\chi^{im\setminus j}_{C,2}$};
      
      \node [factor, right=of x_ijm,   label={$f^{ijm}_{2|1}$}] (f2_ijm) {};
      \node [latent, right=of f2_ijm, xshift=-0.65cm, minimum size=31pt] (x2_ijm) {$\chi^{ijm}_{C,2}$};
      
      \node [factor, right=of x_ij, label={$f^{ij}_{2|1}$}] (f2_ij) {};
      \node [latent, right=of f2_ij, xshift=-0.65cm, minimum size=31pt] (x2_ij) {$\chi^{ij\setminus m}_{C,2}$};
      
      \node [factor,fill=black!40, between=x_Lj and x_ijm] (f_j11) {};
      \node [factor,fill=black!40, between=x_Lj and x_ij] (f_j21) {};
      
      \node [factor,fill=black!40, between=x_Lm and x_ijm] (f_m11) {};
      \node [factor,fill=black!40, between=x_Lm and x_im] (f_m21) {};
      
      
      \node (b) at (5,3) {$(b)$};
      \node (x_Lib)[latent, minimum size=31pt] at (5.25,0) {$\chi^i_L$};
      \node (x_Lmb)[latent,dashed, minimum size=31pt] at (5.9,2.25) {$\chi^m_L$};
      \node (x_Ljb)[latent,dashed, minimum size=31pt] at (5.9,-2.25) {$\chi^j_L$};

      \node [latent, right=of x_Lib, xshift=0.0cm, minimum size=31pt] (x2_ijmb) {$\chi^{ijm}_{C,2}$};
      
      \node [latent, above=of x2_ijmb, yshift=-0.5cm, minimum size=31pt] (x2_imb) {$\chi^{im\setminus j}_{C,2}$};
      
      \node [latent, below=of x2_ijmb, yshift=0.5cm, minimum size=31pt] (x2_ijb) {$\chi^{ij\setminus m}_{C,2}$};
     
     \node [factor,fill=red!100, between=x_Lib and x2_ijmb, label=\textcolor{red!100}{{$f^{i}_2$}} ] (f2b) {};
      
  \end{scope}
  
 \graph {
    (import nodes);
   
    
   
    {x_im,x_Li}--yi11, 
    {x_ijm,x_Li}--yi21,  
    {x_ij,x_Li}--yi31,
    {x_ij,x_ijm}--f_ij,
    {x_im,x_ijm}--f_im,
    {x2_ij,x_ij}--f2_ij,
    {x2_im,x_im}--f2_im,
    {x2_ijm,x_ijm}--f2_ijm,
    {x_Lm}--[dashed]f_m21, 
    {x_Lm}--[dashed]f_m11,
    {x_Lj}--[dashed]f_j21, 
    {x_Lj}--[dashed]f_j11,
    {x_im}--[dashed]f_m21,
    {x_ijm}--[dashed]f_m11,
    {x_ij}--[dashed]f_j21, 
    {x_ijm}--[dashed]f_j11,
    {x_Lib,x2_imb, x2_ijb, x2_ijmb }--f2b,
    {x_Ljb,x_Lmb}--[dashed]f2b,
   
    };
    
\end{tikzpicture}}
\caption{Neighborhood graph perspective: factor graph representing robot $i$' local pdf with hidden local variables of neighboring robots $m$ and $j$. Dashed nodes and grey factors are hidden from robot $i$. (a) graph before marginalization of time step 1 (b) fully connected graph after marginalization. }
      \label{fig:fullGraph}
      \vspace{-0.2in}
\end{figure}

We begin our analysis from time step 1, with the initial pdf $p(\chi^i_{2:1}|Z^{i,+}_1)$ (post fusion) and conditional independence structure of the type shown in Fig. \ref{fig:fullGraph}(a), where $\chi^i_{2:1}$ denotes the augmented state of time steps $1$ and $2$, and $Z^{i,+}_1$ is all the data gathered by robot $i$ up to and including time step 1. The graph also shows `neighborhood variables' not monitored (hidden) from robot $i$ in the dashed nodes and hidden dependencies are noted by the dashed lines. 

It can be seen that non-mutual variables are independent given the common variables, e.g., 
\begin{equation*}
    \chi_L^m\perp \chi_L^i, \chi_{C,1}^{ij\setminus m}, \chi_{C,2}^{ij\setminus m}| \chi_{C,1}^{im}, \ \ \ \chi_{C,1}^{im}=\chi_{C,1}^{ijm}\cup \chi_{C,1}^{im\setminus j}.
\end{equation*}
Here common variables are separated into three different sets $\chi_C^{ijm}$, $\chi_C^{ij\setminus m}$ and $\chi_C^{im\setminus j}$,  respectively representing variables common to the three robots, variables common to $i$ and $j$ but not to $m$, and variables common to $i$ and $m$ but not to $j$. 
Dependencies between these subsets are created by fusion, which introduces the factors $f^{im}_1$ and $f^{ij}_1$, coupling the variables in the common sets $\chi^{im}_{C,1}$ and $\chi^{ij}_{C,1}$, respectively, as a result of marginalization of non-mutual variables at the communicating robots $m$ and $j$, e.g., $p(\chi^{ij}_C|Z^{j,-}_k)$ in (\ref{eq:Heterogeneous_fusion}) \cite{dagan_factor_2021}. A dictionary for the different types of factors and their interpretation is given in Table \ref{tab:factor_definitions}.

\begin{table}[tb]
\renewcommand{\arraystretch}{1.75}
\caption{Factors dictionary, giving examples for different types of factors, their notation and their pdf interpretation.  }
    \begin{center}
    \begin{tabular}{c|c|c}
        Factor  & Type & Proportional to  \\ \hline
        $f^{i,l}_k$ & Local measurement & $p(y^{i,l}_{k}|\chi^{i,l}_{k})$ \\ \hline
        $f^{ij\setminus m}_{2|1}$ & Dynamic prediction   &  $p(\chi^{ij\setminus m}_{C,2}|\chi^{ij\setminus m}_{C,1})$ \\ \hline
        $f_k^{ij}$ & Fusion & $p(\chi^{ij}_C|Z^{j,-}_k)$ (from (\ref{eq:Heterogeneous_fusion}))\\ \hline
         \textcolor{red}{$f^{i}_2$}& Dense marginalization & $\int p(\chi^i_{2:1}|Z^{i,+}_1)d\chi_{C,1}^i$  \\ \hline
         \textcolor{red}{$\tilde{f}^{ij}_2$}& Approximate marginalization & $\int \tilde{p}(\chi^i_{2:1}|Z^{i,+}_1)d\chi_{C,1}^i$ (\ref{eq:local_k21})\\ 
    \end{tabular}
    \end{center}
    \label{tab:factor_definitions}
    \vspace{-0.2in}
\end{table} 

As can be seen from Fig \ref{fig:fullGraph}(a), the graph encodes the following local pdf factorization,
\begin{equation}
    \begin{split}
        &p(\chi^i_{2:1}|Z^{i,+}_1) = p(\chi_L^i)\cdot \\   &p(\chi^{ijm}_{C,1}|\chi_L^i)\cdot p(\chi^{im\setminus j}_{C,1}|\chi^{ijm}_{C,1},\chi_L^i) \cdot p(\chi^{ij\setminus m}_{C,1}|\chi^{ijm}_{C,1},\chi_L^i)\cdot \\
         &p(\chi^{ijm}_{C,2}|\chi^{ijm}_{C,1}) 
        \cdot p(\chi^{im\setminus j}_{C,2}|\chi^{im\setminus j}_{C,1})   \cdot p(\chi^{ij\setminus m}_{C,2}|\chi^{ij\setminus m}_{C,1}), 
    \end{split}
    \label{eq:factorization}
\end{equation}
where the conditioning on the data $Z_1^{i,+}$ is omitted from the right side of the equations for the rest of the paper for brevity. 

In Fig. \ref{fig:fullGraph}(b), marginalization of past common nodes of time step 1 (filtering) results in a dense factor, causing  both `hidden' and `visible' dependencies. Here hidden refers to variables and dependencies being hidden from robot's $i$ perspective (dashed variables and grey factors in Fig. \ref{fig:fullGraph}), and visible suggests the the variables, thus the dependencies, exist in the robot's local graph. 

The problem is therefore formulated in the following way: we seek a method to filtering that: (i) results in a conservative posterior pdf after the next fusion step and (ii) maintains the conditional independence requirement between non-mutual variables. That is, we ask for a conservative approximation $\hat{p}(\cdot)$ of the pdf $p(\cdot)$, 
\begin{equation}
    \begin{split}
        \hat{p}(\chi^i_{2}|Z^{i,+}_1)\succeq \int p(\chi^i_{2:1}|Z^{i,+}_1)d\chi_{C,1}^i,
    \end{split}
    \label{eq:approx_definition}
\end{equation}
where `$\succeq$' denotes conservative. The question regarding the meaning of conservative in the case of general pdfs is beyond the scope of this paper and the reader is referred to \cite{lubold_formal_2021} and \cite{dagan_exact_2021} for further discussion. It will be assumed later that the distributions can be described using their first two moments (mean and covariance), so that then a commonly used definition of conservativeness exists.

We define a conservative posterior pdf as one that does not underestimate the uncertainty over the robot's set of random variables  $\chi^{i}_{k}\subseteq V_k$ relative to the marginal pdf over $\chi^{i}_{k}$ of a consistent centralized pdf. 
Consistency here means that the fused result does not overestimate or underestimate the true uncertainty. 
The centralized pdf refers to the posterior pdf over the full global set of random variables $\chi_k$, conditioned on all the available data from all the robots up and including time step $k$, $p(V_k|\bigcup_{i\in N_r} Z^-_{i,k})$. For a more complete discussion on conservative pdfs and point estimates, see \cite{dagan_exact_2021}.

\section{CONSERVATIVE FILTERING}
\label{Sec:cons_filtering}
Our proposed solution to the conservative filtering problem is inspired by the graph-SLAM literature, where the process of removing edges between nodes is commonly referred to as \emph{sparsification}. In graph-SLAM, removal of the robot's pose nodes induces dependencies between landmarks and results in a dense information matrix. 
Several methods have been proposed to enforce conditional independence between the landmarks in a conservative manner, which differ in the `timing' of the edge removal. For example, the \emph{Exactly Sparse Extended Information Filter} (ESEIF) algorithm \cite{walter_exactly_2007} removes motion links \emph{prior} to marginalization of pose nodes, to induce a sparse marginalized information matrix. On the other hand, the algorithms in \cite{vial_conservative_2011}, \cite{carlevaris-bianco_conservative_2014} enforce conditional independence \emph{after} marginalization by manipulation of the marginalized dense information matrix. Next, we make use of these two concepts to derive a new conservative filtering algorithm. We divide the problem into two parts, first to account for the hidden dependencies between local variables, and second to enforce conditional independence between sets of visible non-mutual variables.

\subsection{Local Variables - Hidden Dependencies}
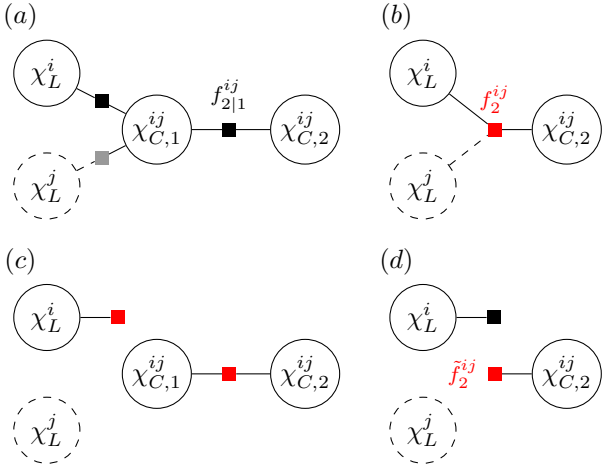
\begin{figure}[tb]
\begin{tikzpicture}[ new set=import nodes]
 \begin{scope}[nodes={set=import nodes}]
      
      \node (a) at (-0.35,1.5) {$(a)$};
      \node (s1)[latent, minimum size=25pt] at (0,0.75) {$\chi_L^i$};
      \node (s2)[latent, dashed, minimum size=25pt] at (0,-0.75) {$\chi_L^j$};
      \node [latent, right=of s1, yshift=-0.75cm, xshift=-0.45cm] (x1) {$\chi_{C,1}^{ij}$};
      \node [factor, right=of x1, label={$f^{ij}_{2|1}$}] (f1) {};
      \node [latent, right=of f1, xshift=-0.55cm] (x2) {$\chi_{C,2}^{ij}$};
      \node [factor, between=x1 and s1 ] (yi1) {};
      \node [factor,fill=black!40, between=x1 and s2 ] (yj1) {};
      
      \node (b) at (4.65,1.5) {$(b)$};
      \node (s3)[latent, minimum size=25pt] at (5,0.75) {$\chi_L^i$};
      \node (s4)[latent, dashed, minimum size=25pt] at (5,-0.75) {$\chi_L^j$};
      \node [latent, right=of s3, yshift=-0.75cm] (x3) {$\chi_{C,2}^{ij}$};
      \node [factor,fill=red!100, between=x3 and s3, yshift=-0.375cm, label=\textcolor{red!100}{$f^{ij}_2$}]  (f2) {};
     
      \node (c) at (-0.35,-1.75) {$(c)$};
      \node (s1c)[latent, minimum size=25pt] at (0,-2.5) {$\chi_L^i$};
      \node (s2c)[latent, dashed, minimum size=25pt] at (0,-4) {$\chi_L^j$};
      \node [latent, right=of s1c, yshift=-0.75cm, minimum size=25pt, xshift=-0.45cm] (x1c) {$\chi_{C,1}^{ij}$};
      \node [factor, right=of x1c, fill=red!100 ] (f1c) {};
      \node [latent, right=of f1c, xshift=-0.55cm, minimum size=25pt] (x2c) {$\chi_{C,2}^{ij}$};
      \node [factor, right=of s1c, fill=red!100 ] (yi1siC) {};
      
      \node (d) at (4.65,-1.75) {$(d)$};
      \node (s1d)[latent, minimum size=25pt] at (5,-2.5) {$\chi_L^i$};
      \node (s2d)[latent, dashed, minimum size=25pt] at (5,-4) {$\chi_L^j$};
      \node [latent, right=of s1d, yshift=-0.75cm, minimum size=25pt] (x2d) {$\chi_{C,2}^{ij}$};
      \node [factor,fill=red!100, between=x2d and s1d, yshift=-0.375cm, label=left:\textcolor{red!100}{{$\tilde{f}^{ij}_2$}}]  (f2d) {};
      \node [factor, right=of s1d] (yi1siD) {};
     
  \end{scope}
  
 \graph {
    (import nodes);
   
    x1--x2, 
    {x1,s1}--yi1, {s2}--[dashed]yj1, {x1}--yj1,
    {s3,x3}--f2, s4--[dashed]f2,
   
    {x1c,x2c}--f1c, 
    s1c--yi1siC,  
    
    x2d--f2d, 
    s1d--yi1siD,
    };
    
\end{tikzpicture}
\caption{Example of hidden correlation due to filtering, demonstrated on robot $i$'s local graph, only shown factors relevant to the example. Dashed variable nodes and edges are hidden from robot $i$. (a) robot $i$'s graph after first measurement, coupling local and common variables, fusion and prediction step. (b) filtering, with the marginalization of $\chi_{C,1}^{ij}$ results in coupling shown by the red factor, which is hidden from the robot (as $i$ has no knowledge about $\chi_L^j$) but evident in the full graph. (c)-(d) demonstrates the proposed solution, approximating by marginal distributions prior to filtering.  }
      \label{fig:local_variables}
      \vspace{-0.2in}
\end{figure}

To understand the problem of hidden dependencies between local variables it is sufficient to study the reduced problem of only two robots. Consider two robots $i$ and $j$, with local subsets of variables $\chi_L^i$ and $\chi_L^j$ and a common subset $\chi_C^{ij}$. 
Fig. \ref{fig:local_variables}(a)-(b) shows an example of hidden correlation due to filtering, demonstrated on robot $i$'s local graph. Robot $i$ has no knowledge on robot $j$'s local variables, which now have hidden dependencies with its local variables due to marginalization. To avoid this hidden coupling robot $i$ needs to take preventive action prior to marginalization based on its local understanding of the distribution. 

The local pdf at robot $i$, prior to marginalization (Fig. \ref{fig:local_variables}(a)), can be factorized as,
\begin{equation}
    \begin{split}
        p(\chi^i_{2:1}|Z^{i,+}_1) = p(\chi_{C,1}^{ij},\chi_{C,2}^{ij})\cdot p(\chi^i_L|\chi_{C,1}^{ij}).
    \end{split}
    \label{eq:local_exact_eq}
\end{equation}
By removing the dependency between the local and common variables prior to marginalization, i.e., approximating the pdf by its marginals (Fig. \ref{fig:local_variables}(c)),
\begin{equation}
    \begin{split}
        \tilde{p}(\chi^i_{2:1}|Z^{i,+}_1) = p(\chi_{C,1}^{ij},\chi_{C,2}^{ij})\cdot p(\chi^i_L),
    \end{split}
    \label{eq:local_k21}
\end{equation}
and only then marginalize out $\chi_{C,1}^{ij}$ (Fig. \ref{fig:local_variables}(d)),
\begin{equation}
    \begin{split}
        \tilde{p}(\chi^i_{2}|Z^{i,+}_1) = p(\chi_{C,2}^{ij})\cdot p(\chi^i_L),
    \end{split}
    \label{eq:local_k1}
\end{equation}
the hidden dependencies are prevented.
Notice that from robot $i$'s perspective, it didn't have to `know' if robot $j$ is tasked with more local variables, but rather it just assumes that there are other variables that are not known locally and takes preemptive measures to avoid hidden dependencies. 

\subsection{Common Variables - Visible Dependencies}
\begin{figure}[tb]
\begin{tikzpicture}[ new set=import nodes]
 \begin{scope}[nodes={set=import nodes}]

      \node (a) at (-0.35,3) {$(a)$};
      \node (x_Lia)[latent, minimum size=31pt] at (0,0) {$\chi^i_L$};
      \node (x_Lma)[latent,dashed, minimum size=31pt] at (0.5,2.25) {$\chi^m_L$};
      \node (x_Lja)[latent,dashed, minimum size=31pt] at (0.5,-2.25) {$\chi^j_L$};

      \node [latent, right=of x_Lia, xshift=0.0cm, minimum size=31pt] (x2_ijma) {$\chi^{ijm}_{C,2}$};
      
      \node [latent, above=of x2_ijma, yshift=-0.5cm, minimum size=31pt] (x2_ima) {$\chi^{im\setminus j}_{C,2}$};
      
      \node [latent, below=of x2_ijma, yshift=0.5cm, minimum size=31pt] (x2_ija) {$\chi^{ij\setminus m}_{C,2}$};
     
     \node [factor,fill=red!100, right=of x2_ijma ] (f2a) {};
     \node [factor,fill=red!100, above=of x_Lia ] (fla) {};
     
      \node (b) at (3.85,3) {$(b)$};
      \node (x_Lib)[latent, minimum size=31pt] at (4.2,0) {$\chi^i_L$};
      \node (x_Lmb)[latent,dashed, minimum size=31pt] at (4.7,2.25) {$\chi^m_L$};
      \node (x_Ljb)[latent,dashed, minimum size=31pt] at (4.7,-2.25) {$\chi^j_L$};

      \node [latent, right=of x_Lib, xshift=0.0cm, minimum size=31pt] (x2_ijmb) {$\chi^{ijm}_{C,2}$};
      
      \node [latent, above=of x2_ijmb, yshift=-0.5cm, minimum size=31pt] (x2_imb) {$\chi^{im\setminus j}_{C,2}$};
      
      \node [latent, below=of x2_ijmb, yshift=0.5cm, minimum size=31pt] (x2_ijb) {$\chi^{ij\setminus m}_{C,2}$};
     
     \node [factor,fill=red!100, right=of x2_ijmb ] (f2b) {};
     \node [factor,fill=red!100, above=of x_Lib ] (flb) {};
     
     \node [factor,fill=black!40, between=x_Ljb and x2_ijmb] (f_j11) {};
     \node [factor,fill=black!40, between=x_Ljb and x2_ijb] (f_j21) {};
      
     \node [factor,fill=black!40, between=x_Lmb and x2_ijmb] (f_m11) {};
     \node [factor,fill=black!40, between=x_Lmb and x2_imb] (f_m21) {};
     
     \node [factor, between=x_Lib and x2_imb ] (yi11) {};
     \node [factor, between=x_Lib and x2_ijmb ] (yi21) {};
     \node [factor, between=x_Lib and x2_ijb ] (yi31) {};
     
      
      \node (c) at (-0.35,-3.25) {$(c)$};
      \node (x_Lic)[latent, minimum size=31pt] at (0,-6.25) {$\chi^i_L$};
      \node (x_Lmc)[latent,dashed, minimum size=31pt] at (0.5,-4) {$\chi^m_L$};
      \node (x_Ljc)[latent,dashed, minimum size=31pt] at (0.5,-8.5) {$\chi^j_L$};

      \node [latent, right=of x_Lic, xshift=0.0cm, minimum size=31pt] (x2_ijmc) {$\chi^{ijm}_{C,2}$};
      
      \node [latent, above=of x2_ijmc, yshift=-0.5cm, minimum size=31pt] (x2_imc) {$\chi^{im\setminus j}_{C,2}$};
      
      \node [latent, below=of x2_ijmc, yshift=0.5cm, minimum size=31pt] (x2_ijc) {$\chi^{ij\setminus m}_{C,2}$};
     
     \node [factor,fill=red!100, between=x2_ijmc and x2_imc, label=right:\textcolor{red!100}{{$\tilde{f}^{im}_{2}$}} ] (f2c_im) {};
     \node [factor,fill=red!100, between=x2_ijmc and x2_ijc, label=right:\textcolor{red!100}{{$\tilde{f}^{ij}_{2}$}} ] (f2c_ij) {};
     
     \node [factor,fill=red!100, above=of x_Lic ] (flc) {};
     
      \node (d) at (3.85,-3.25) {$(d)$};
      \node (x_Lid)[latent, minimum size=31pt] at (4.2,-6.25) {$\chi^i_L$};
      \node (x_Lmd)[latent,dashed, minimum size=31pt] at (4.7,-4) {$\chi^m_L$};
      \node (x_Ljd)[latent,dashed, minimum size=31pt] at (4.7,-8.5) {$\chi^j_L$};

      \node [latent, right=of x_Lid, xshift=0.0cm, minimum size=31pt] (x2_ijmd) {$\chi^{ijm}_{C,2}$};
      
      \node [latent, above=of x2_ijmd, yshift=-0.5cm, minimum size=31pt] (x2_imd) {$\chi^{im\setminus j}_{C,2}$};
      
      \node [latent, below=of x2_ijmd, yshift=0.5cm, minimum size=31pt] (x2_ijd) {$\chi^{ij\setminus m}_{C,2}$};
     
     \node [factor,fill=red!100, between=x2_ijmd and x2_imd, label=right:\textcolor{red!100}{{$\tilde{f}^{im}_{2}$}} ] (f2d_im) {};
     \node [factor,fill=red!100, between=x2_ijmd and x2_ijd, label=right:\textcolor{red!100}{{$\tilde{f}^{ij}_{2}$}} ] (f2d_ij) {};
     
     \node [factor,fill=red!100, above=of x_Lid ] (fld) {};
     
     \node [factor,fill=black!40, between=x_Ljd and x2_ijmd] (f_j11d) {};
     \node [factor,fill=black!40, between=x_Ljd and x2_ijd] (f_j21d) {};
      
     \node [factor,fill=black!40, between=x_Lmd and x2_ijmd] (f_m11d) {};
     \node [factor,fill=black!40, between=x_Lmd and x2_imd] (f_m21d) {};
     
     \node [factor, between=x_Lid and x2_imd ] (yi11d) {};
     \node [factor, between=x_Lid and x2_ijmd ] (yi21d) {};
     \node [factor, between=x_Lid and x2_ijd ] (yi31d) {};
      
  \end{scope}
  
 \graph {
    (import nodes);
   
    
   
    
    {x2_ima, x2_ija, x2_ijma }--f2a,
    {x_Lia }--fla,
    
    {x_Lib,x2_ijmb}--yi21,
    {x_Lib,x2_imb}--yi11,
    {x_Lib,x2_ijb}--yi31,
    {x2_imb, x2_ijb, x2_ijmb }--f2b,
    {x_Lmb}--[dashed]f_m21, 
    {x_Lmb}--[dashed]f_m11,
    {x_Ljb}--[dashed]f_j21, 
    {x_Ljb}--[dashed]f_j11,
    {x2_imb}--[dashed]f_m21,
    {x2_ijmb}--[dashed]f_m11,
    {x2_ijb}--[dashed]f_j21, 
    {x2_ijmb}--[dashed]f_j11,
    {x_Lib }--flb,

    {x2_ijmc, x2_imc}--f2c_im,
    {x2_ijmc, x2_ijc}--f2c_ij,
    {x_Lic }--flc,
    
    {x_Lid,x2_ijmd}--yi21d,
    {x_Lid,x2_imd}--yi11d,
    {x_Lid,x2_ijd}--yi31d,
    {x_Lid }--fld,
  
    {x_Lmd}--[dashed]f_m21d, 
    {x_Lmd}--[dashed]f_m11d,
    {x_Ljd}--[dashed]f_j21d, 
    {x_Ljd}--[dashed]f_j11d,
    {x2_imd}--[dashed]f_m21d,
    {x2_ijmd}--[dashed]f_m11d,
    {x2_ijd}--[dashed]f_j21d, 
    {x2_ijmd}--[dashed]f_j11d,
    {x2_ijmd, x2_imd}--f2d_im,
    {x2_ijmd, x2_ijd}--f2d_ij,
    
    };
    
\end{tikzpicture}
\caption{Visible dependencies with full graph perspective. (a) graph after filtering (b) addition of measurement factors and dependencies at time step 2, red dense factor breaks conditional independence assumption required for heterogeneous fusion. (c)-(d) our proposed method to regain conditional independence by factorizing into smaller local factors.}
      \label{fig:common_vars}
      \vspace{-0.2in}
\end{figure}
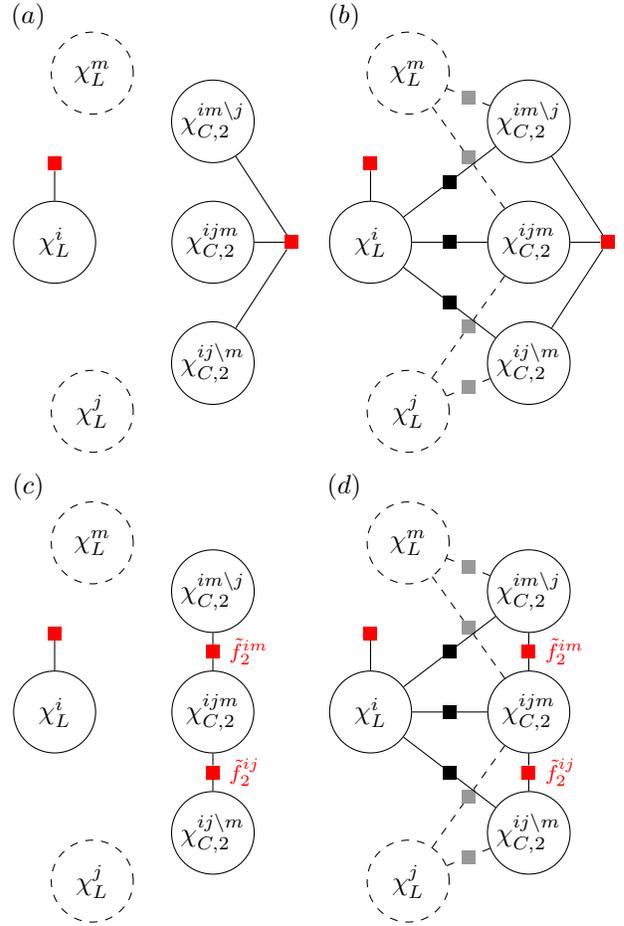

To analyze the visible dependencies between the different sets of common variables, consider again the 3 robots $j-i-m$ described in Sec. \ref{Sec:prob_statement} and assume that edges to the local variables have been removed prior to marginalization (\ref{eq:local_k21})-(\ref{eq:local_k1}). Thus the current factor graph at robot $i$ has the form depicted in Fig. \ref{fig:common_vars}(a) and it describes the approximate pdf,
\begin{equation}
    \begin{split}
        \tilde{p}(\chi^i_{2}|Z^{i,+}_1) = p(\chi_L^i)\cdot    p(\chi^{ijm}_{C,2}, \chi^{im\setminus j}_{C,2}, \chi^{ij\setminus m}_{C,2}).
    \end{split}
    \label{eq:common_exact_eq}
\end{equation}

The robots now take a second measurement shown by the black and grey factors in Fig. \ref{fig:common_vars}(b). Here the grey factors show measurements taken by robots $j$ and $m$, which are not available at robot $i$ prior to the next fusion episode. It can be seen from the factor graph (Fig. \ref{fig:common_vars}(b)) that the conditional independence assumption between non-mutual variables given common variables does not hold, i.e., $\chi_L^j \not\perp \chi_{C,2}^{im\setminus j}\cup \chi_{L}^i|\chi_C^{ij}$ for robots $i$ and $j$ and similarly $\chi_L^m \not\perp \chi_{C,2}^{ij\setminus m}\cup \chi_{L}^i|\chi_C^{im}$ for robots $i$ and $m$. This breaks the assumption allowing heterogeneous fusion according to (\ref{eq:Heterogeneous_fusion}) and might cause overconfidence in robot $i$'s estimate. 

Fig. \ref{fig:common_vars}(d) shows that by requiring the inner-structure/dependencies between the subsets of common variables at time step 2 to be identical to the dependencies of those subsets at time step 1 prior to marginalization, the conditional independence required for heterogeneous fusion is recovered. For example, if $\chi_{C,1}^{im\setminus j}\perp \chi_{C,1}^{ij\setminus m} | \chi_{C,1}^{ijm}, \chi_L^i$ prior to marginalization (Fig. \ref{fig:fullGraph}(a)), we require $\chi_{C,2}^{im\setminus j}\perp \chi_{C,2}^{ij\setminus m} | \chi_{C,2}^{ijm}, \chi_L^i$ post marginalization (Fig. \ref{fig:common_vars}(d)).   
This requirement also bypasses the need for `bookkeeping' of the different subset of variables, as it simply keeps the same structure as before filtering. 

The method to enforce this structure is by factorizing the dense factor induced by marginalization shown in Fig. \ref{fig:common_vars}(a) to two smaller factors $\tilde{f}^{im}_2(\chi_C^{im})$ and $\tilde{f}^{ij}_2(\chi_C^{ij})$, connecting separately the variables common to $i$ and $m$ and $i$ and $j$, respectively. In other words, we wish to further approximate the pdf $\tilde{p}(\chi^i_{2}|Z^{i,+}_1)$ shown in Fig. \ref{fig:common_vars}(a) with the pdf $\hat{p}(\chi^i_{2}|Z^{i,+}_1)$ shown in Fig. \ref{fig:common_vars}(c), such that:
\begin{equation}
    \begin{split}
        &\hat{p}(\chi^i_{2}|Z^{i,+}_1)=\\
        &p(\chi_L^i)\cdot p(\chi_{C,2}^{ijm})\cdot 
        p(\chi_{C,2}^{ij\setminus m}|\chi_{C,2}^{ijm})\cdot p(\chi_{C,2}^{im\setminus j}|\chi_{C,2}^{ijm}).
    \end{split}
    \label{eq:final_approximation}
\end{equation}
The steps taken thus far have led to an approximate pdf that accounts for the hidden and visible dependencies in the data in a way that preserves the conditional independence assumptions. However, we have yet to guarantee the conservativeness of the approximate pdf with respect to (w.r.t) the true pdf as required in the problem statement (\ref{eq:approx_definition}), where true pdf here means the dense pdf that would have been resulted without the approximations. While no assumption on the type of distribution have been made until now, there is no commonly used formal definition of conservativeness for general pdfs. Thus we now focus our attention to the case where the pdf can be represented by a Gaussian distribution (or the first two moments). These are used in many applications across robotics \cite{dellaert_factor_2021}.

Assume that the factors are Gaussian functions, described by their information matrix (inverse of covariance). Then the factorization, or sparsification given in (\ref{eq:final_approximation}), can be preformed using various methods, e.g., \cite{vial_conservative_2011}, \cite{carlevaris-bianco_conservative_2014}, \cite{forsling_consistent_2019}. The basic idea of all of these methods is the similar: given the true Gaussian distribution, shown in Fig. \ref{fig:fullGraph}(b) and characterized by its \emph{dense} information matrix and vector $\mathcal{N}(\zeta_{tr},\Lambda_{tr})$. We wish to make the approximate sparse Gaussian pdf, $\hat{p}(\chi^i_{2}|Z^{i,+}_1)\sim \mathcal{N}(\zeta_{sp},\Lambda_{sp})$, shown in Fig. \ref{fig:common_vars}(c) and given in (\ref{eq:final_approximation}), conservative in the positive semi-definite (PSD) sense,
\begin{equation}
    \begin{split}
        \Lambda_{tr}-\Lambda_{sp}\succeq 0.
    \end{split}
    \label{eq:consFilter}
\end{equation}

Due to its relative simplicity and the fact that it does not require optimization, we choose to use the method suggested by \cite{forsling_consistent_2019} and generalized in \cite{dagan_exact_2021}. Briefly, we solve for the deflation constant $\lambda_{min}$, the minimal eigenvalue of $\Tilde{Q}=\Lambda_{sp}^{-\frac{1}{2}}\Lambda_{tr}\Lambda_{sp}^{-\frac{1}{2}}$ and enforce the mean of the sparse distribution to equal the true one. The conservative approximate pdf is,
\begin{equation}
    \begin{split}
        \hat{p}(\chi^i_{2}|Z^{i,+}_1)\sim \mathcal{N}(\lambda_{min}\Lambda_{sp}\Lambda_{tr}^{-1}\zeta_{tr}, \lambda_{min}\Lambda_{sp}).
    \end{split}
    \label{eq:sparse_cons_mat}
\end{equation}
Note that to apply the deflation back to the factor graph expressing this pdf, factors need to be manipulated. In the case of Gaussian distributions, the information matrix, like the corresponding factor graph, directly expresses the conditional independence structure. Thus zero terms in the information matrix indicate conditional independence (lack of factor) between the variables, and non-zero terms correspond to a factor between variables. We then re-factorize the graph in Fig. \ref{fig:common_vars}(c) in the following way:\footnote{Since the information form is related to the log of the distribution, it can be expressed as summation instead of multiplication \cite{koller_probabilistic_2009}}.
\begin{equation}
    \begin{split}
        \hat{p}(\chi^i_{2}|Z^{i,+}_1)\propto &f(\chi_L^i)+f(\chi_{C,2}^{im\setminus j})+f(\chi_{C,2}^{ijm})+f(\chi_{C,2}^{ij\setminus m})+\\
        &f(\chi_{C,2}^{ijm},\chi_{C,2}^{im\setminus j})+f(\chi_{C,2}^{ijm},\chi_{C,2}^{ij\setminus m}),
    \end{split}
    \label{eq:refactorization}
\end{equation}
where factors over one set of variables, e.g., $f(\chi_{C,2}^{ijm})$, are Gaussians with the vector and block diagonal elements of the information vector and matrix in (\ref{eq:sparse_cons_mat}) corresponding to the variables in the set. Factors over two sets of variables hold a zero information vector and the matrix of off-block diagonal elements of those sets are zero block-diagonal elements, e.g., 
\begin{equation*}
    \begin{split}
        &f(\chi_{C,2}^{ijm})=\mathcal{N}(\zeta_{\chi_{C,2}^{ijm}}, \Lambda_{\chi_{C,2}^{ijm},\chi_{C,2}^{ijm}}), \\ 
        &f(\chi_{C,2}^{ijm},\chi_{C,2}^{im\setminus j}) = \mathcal{N}(\begin{bmatrix} \underline{0}  \\ \underline{0}  \end{bmatrix},
        \begin{bmatrix}
        \underline{\underline{0}}  & \Lambda_{\chi_{C,2}^{ijm},\chi_{C,2}^{im\setminus j}}\\ \Lambda_{\chi_{C,2}^{ijm},\chi_{C,2}^{im\setminus j}}^T  & \underline{\underline{0}} \end{bmatrix}).
    \end{split}
\end{equation*}
Here $\underline{0}$ and \underline{\underline{0}} denotes the zero vector and matrix in the correct dimensions, respectively.

\subsection{Updating the CF}
The last step of the proposed method is to guarantee that data is not over-discounted in the fusion step. Remember that on every communication channel a robot has with its neighbors, a CF was added to track dependencies in the data (Sec. \ref{subsec:CF}), in the form of another factor graph over the common variables. The conservative filtering operations so far did not affect the CF while discounting information on the robot's graph. This might lead to a situation where the CF has `more information', in the sense of information matrix, over the common variables than the robot itself. During fusion this might result robot $i$ sending `negative information' to its neighbors. To avoid this situation, we update every CF graph the robot maintains by deflating the factors of the graph with the deflation constant $\lambda_{min}$. In practice that means multiplying the information vector and matrix of every factor $f$ in the CF graph by $\lambda_{min}$.     

\subsection{Algorithm}
The full conservative filtering approach is now summarized in Algorithm \ref{algo:cons_filter}, which is written from the perspective of one robot $i$ communicating with its $n^i_r$ neighbors and is applied recursively at every filtering step.
Note that while the algorithm defines operations on a factor graph and in the example of this paper is integrated with the FG-DDF framework, it is not limited to a specific type of graphical model nor a pdf or a fusion algorithm. The main steps, given in lines 4-7, can in general, be defined by various methods, with the only caveat being guaranteeing conservativeness for non-Gaussian pdfs, which, in practice is frequently solved by approximating with a Gaussian as done in many robotic applications.

\begin{algorithm}[h!]
    \caption{Conservative Filtering}
    \label{algo:cons_filter}
    \begin{algorithmic}[1]
    \State \textbf{Input:} Local factor graph $\mathcal{F}^{i}$
    \State Prediction step
    \State Create a copy of the `true' graph $\mathcal{F}^{i}_{tr}$ 
    \State Approximate $\mathcal{F}^{i}$ with marginal pdfs \Comment{{\color{Gray} Eq. (\ref{eq:local_k21}) }}
    \State Marginalize out past nodes in $\mathcal{F}^{i}$ (\ref{eq:local_k1}) and $\mathcal{F}^{i}_{tr}$
    \State Regain conditional independence in $\mathcal{F}^{i}$ \Comment{{\color{Gray} Eq. (\ref{eq:final_approximation}) }}
    \State Guarantee conservativeness w.r.t  $\mathcal{F}^{i}_{tr}$ \Comment{{\color{Gray} Eq. (\ref{eq:sparse_cons_mat}) }}
    \State Re-factorize graph \Comment{{\color{Gray} Eq. (\ref{eq:refactorization}) }}
    \For{Every neighbor $j\in N_r^i$ }
    \State Update CF graph $\mathcal{F}^{ij}_{CF}$
    \EndFor
    \State \Return
    \end{algorithmic}
\end{algorithm}

\section{SIMULATION RESULTS and DISCUSSION}
\label{Sec:sim}
To test the proposed method, a simulation of 4 robots tracking 5 dynamic targets was preformed. The robots are connected in a chain  
$(1\leftrightarrow 2\leftrightarrow 3\leftrightarrow 4)$ with bidirectional communication. Each robot is tasked with estimating the $2D$ (north, east) position and velocity $x^t=[n^t,\dot{n}^t,e^t,\dot{e}^t]^T$ of a subset of the 5 targets, and its own constant (but unknown) robot-to-target relative position measurement bias $s^i=[b^i_{n},b^i_{e}]^T$, similar to the bias in \cite{noack_treatment_2015}. Robots tasks, described by their local random state vector, are as follows:
\begin{equation}
    \begin{split}
        \chi^1_k=\begin{bmatrix} x^1_k \\ x^2_k \\ s^1 \end{bmatrix},
        \chi^2_k=\begin{bmatrix} x^2_k \\ x^3_k \\ s^2 \end{bmatrix},
        \chi^3_k=\begin{bmatrix} x^3_k \\ x^4_k \\ x^5_k \\ s^3 \end{bmatrix},
        \chi^4_K=\begin{bmatrix} x^4_k \\ x^5_k \\ s^4 \end{bmatrix}.
    \end{split}
\end{equation}
For example, robot 1 is tasked with tracking targets 1 and 2 and its own local bias. Common states between robots are those states in the intersection between the state vectors, e.g, $\chi^{12}_{C,k} = x^2_k$ is the common state vector between robots 1 and 2, and $\chi^1_{L,k}=[(x^1_k)^T, (s^1)^T]^T$ are the local state vectors at robot 1. Notice that in homogeneous fusion, all 4 robots reason and communicate the full global state, including 28 states. On the other hand, in heterogeneous fusion robots only reason over their local tasks, including maximum 14 states (robot 3) and communicate over maximum 8 common states (robots 3-4). For Gaussian distributions that translates to more than $95\%$ reduction in communication and computation costs for robots 1,2,4 and about $90\%$ for robot 3.

In every time step $k$, each robot $i$ takes two types of measurements, a relative measurement to target $t$, $y^{i,t}_{k}$,  and a measurement to a known landmark, $m^i_{k}$,
\begin{equation}
    \begin{split}
        &y^{i,t}_{k} = x^t_k+s^i+v^{i,1}_k, \ \ v^{i,1}_k \sim \mathcal{N}(0,R^{i,1}_k),  \\
        &m^i_{k} = s^i+v^{i,2}_k, \ \ v^{i,2}_k \sim \mathcal{N}(0,R^{i,2}_k).
    \end{split}
    \label{eq:meas_model}
\end{equation}

As in many target tracking problems \cite{bar-shalom_linear_2001}, each target $t$ dynamics is model by the following kinematic model,
\begin{equation}
    \begin{split}
    &x^t_{k+1}=Fx^t_{k}+Gu^t_k+\omega_k, \ \ \omega_k \sim \mathcal{N}(0,0.08\cdot I_{n_x\times n_x}),\\
    &F=\begin{bmatrix}1 & \Delta t &0 &0\\0 &1 &0 &0\\ 0 &0 &1 & \Delta t\\0& 0 &0 &1 \end{bmatrix}, \quad
    G=\begin{bmatrix}\frac{1}{2}\Delta t^2 &0\\\Delta t&0\\0 &\frac{1}{2}\Delta t^2\\0 &\Delta t \end{bmatrix}.
    \end{split}
    \label{eq:dynamicEq}
\end{equation}

The simulation uses the FG-DDF algorithm \cite{dagan_factor_2021}, with the addition of the conservative filtering algorithm \ref{algo:cons_filter}. Each robot then maintains and reasons over its local dynamic factor graph and estimates the current MMSE estimate (mean and covariance) using the sum-product (message passing) algorithm on the factor graph \cite{frey_factor_1997}.

\begin{figure}[bt]
	\centering
     \includegraphics[width=0.47\textwidth]{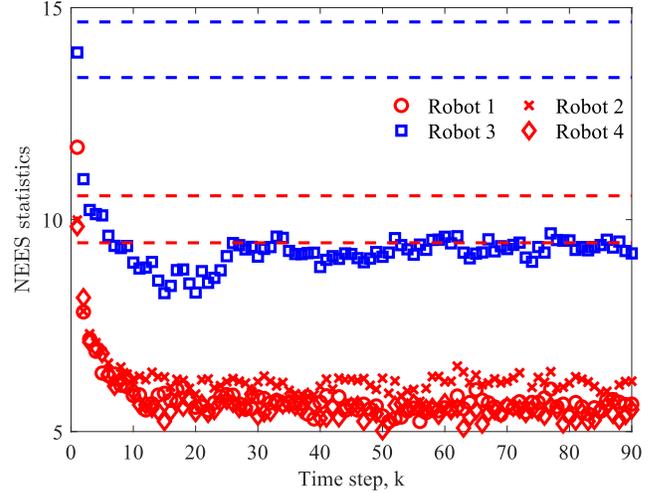}
	\caption{NEES test results from 250 Monte Carlo of 4-robots, 5-targets tracking simulations. Shown are $95\%$ confidence bounds.}
	\label{fig:NEES_simResults}
	\vspace{-0.2in}
\end{figure}

The algorithm's performance was tested with 250 Monte Carlo simulations. Consistency of each robot's estimate was tested using the normalized estimation error squared (NEES) test \cite{bar-shalom_linear_2001}.  Fig. \ref{fig:NEES_simResults} shows the four robots' results with $95\%$ confidence bounds for robots 1,2,4 in dashed red centered around 10 (number of estimated states) and for robot 3 in dashed blue around 14. Results show that all robots are consistent, underestimating their uncertainty due to due to the information matrix deflation (covariance inflation) in the conservative filtering process.

\begin{figure}[bt]
	\centering
     \includegraphics[width=0.48\textwidth]{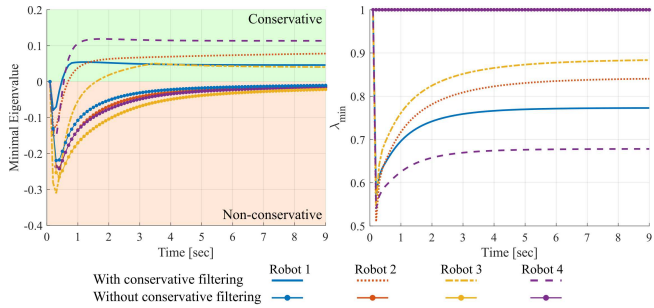}
	\caption{Minimal eigenvalue of the covariance difference (left) and deflation constant (right) of the four robots. Shown are results with and without the conservative filtering algorithm.   }
	\label{fig:lamda}
	\vspace{-0.2in}
\end{figure}

To validate that the robots' estimates are conservative, the local uncertainty of each robot was compared with a centralized estimator over the full 28 random state vector, getting all data from all the robots at every time step. We thus require the robot's covariance to be larger than the centralized covariance in the PSD sense, $\Sigma_{\chi^i}-\Sigma_{\chi^i}^{cent} \succeq 0$, where $\Sigma_{\chi^i}^{cent}$ is the marginal covariance over the local random state vector of robot $i$, $\chi^i$, taken from the joint centralized covariance over the full system state $\chi$. In practice, this is verified by computing the minimal eigenvalue of the above covariance difference and testing that it is not negative. The left plot of Fig. \ref{fig:lamda} shows the minimal eigenvalue of the covariance difference with and without the conservative filtering algorithm. It can be seen that without the algorithm the heterogeneous fusion results are not conservative. On the other hand, with the algorithm 
it takes about $1-1.5$ seconds for the robots' estimates to become conservative, but once it is conservative, it stays conservative. 

The right plot of Fig. \ref{fig:lamda} shows the change in the deflation constant in time across all robots. Intuitively, as robots accumulate more data, approximations by detaching dependencies have lower impact on the pdf, i.e., the sparse approximation is `closer' to the dense pdf. Thus the deflation constant approaches to some limit, depending on the problem statistics and structure. Also, notice that robots 1 and 4, which only communicate with one robot each, and thus accumulate less data, need smaller constants to regain conservativeness.  

Another interesting observation is that both the minimal eigenvalues and deflation constant are constant across simulations. We attribute it to the fact that these are functions of the problem statistics, i.e., measurement and dynamic model noise, and of the communication network, all where kept constant across simulations. Thus in general, these can be analyzed and studied priori, with no connection to the actual measurements received.   

\section{CONCLUSIONS}
\label{Sec:conclusions}
Heterogeneous Bayesian DDF enables a system of robots to cooperatively share information in a scalable way, such that every robot only reasons about its variables of interest. These can include locally observed variables or any other subset of the global set of variables. By reducing their local set and communicating only common variables of interest, robots can significantly reduce their communication and computation costs.

This paper rigorously develops the theory for conservative filtering in decentralized multi-robot dynamic applications. The analysis and methods are developed in a Bayesian framework using factor graphs, representing each robot's local pdf. When the notion of conservativeness can be defined, e.g., for Gaussian distributions, a practical algorithm is developed and it is shown to yield a conservative estimate for a multi-robot multi-target tracking application. The methods developed are based on only the local knowledge a robot has regarding the structure of the pdf. Thus, robots can independently reason about their local tasks without the need for some global graph knowledge, and opportunistically fuse data for cooperative robots to obtain conservative state estimates. 

The suggested conservative filtering algorithm bridges the gap found in \cite{dagan_exact_2021}, and allows, together with the FG-DDF framework \cite{dagan_factor_2021} to solve different robotic problems, such as target tracking \cite{dagan_exact_2021} and terrain height mapping \cite{schoenberg_distributed_2009}.

Future research will explore non-linear problems, hardware implementation and cyclic network topologies.

\addtolength{\textheight}{-12cm}   






\bibliographystyle{IEEEtran}
\bibliography{references.bib}

\end{document}